\documentclass[11pt]{article}
\usepackage{coling2016}
\usepackage{times}
\usepackage{url}
\usepackage{latexsym}

\usepackage{amsmath}
\usepackage{multirow}
\usepackage{url}
\usepackage{graphics}
\usepackage{graphicx}
\usepackage{epsfig}
\usepackage{subfig}
\usepackage{enumerate}

\title{A Joint Framework for Argumentative Text Analysis Incorporating Domain Knowledge}

\author{
\\ \textbf{Zhongyu Wei$^{1}\thanks{*Corresponding Author: zywei@fudan.edu.cn}$},  \textbf{Chen Li$^{2}$}, \textbf{Yang Liu$^{3}$}\\
$^1$School of Data Science, Fudan University, Shanghai, P.R.China\\
$^2$ Microsoft, Bellevue, WA, USA\\
$^3$Computer Science Department, The University of Texas at Dallas\\
Richardson, Texas 75080, USA\\
 }

\date{}

\begin{document}
\maketitle
\begin{abstract}
For argumentation mining, there are several sub-tasks such as argumentation component type classification, relation classification. Existing research tends to solve such sub-tasks separately, but ignore the close relation between them. In this paper, we present a joint framework incorporating logical relation between sub-tasks to improve the performance of argumentation structure generation. We design an objective function to combine the predictions from individual models for each sub-task and solve the problem with some constraints constructed from background knowledge. We evaluate our proposed model on two public corpora and the experiment results show that our model can outperform the baseline that uses a separate model significantly for each sub-task. Our model also shows advantages on component-related sub-tasks compared to a state-of-the-art joint model based on the evidence graph. \end{abstract}

\section{Introduction}

Argumentation mining has attracted increasing attention from NLP research in recent years. It aims to automatically recognize the structure of argumentation in a text by identifying the type of argumentative discourse unit (ADU, e.g., claim, premises, etc.) and detecting relationships between each pair of such ADUs. A variety of applications can benefit from analyzing argumentative structure of text, including the retrieval of relevant court decisions from legal databases~\cite{palau2009argumentation}, automatic document summarization systems, analysis of scientific papers, and essay scoring~\cite{beigman2014applying,Persing+Ng:15a}. 

The full-fledged task of argumentation mining consists of several sub-tasks including segmentation, identification of ADUs, ADU type classification and relation identification~\cite{peldszus2015joint}. Stab and Gurevych~\shortcite{stab2014identifying} aimed to classify text segments into four classes, namely major claim, claim, premise and non-argumentative for persuasive essays. Based on the same corpus and setting, Nguyen et al.~\shortcite{nguyen2015extracting} explored a semi-supervised method for segment type classification. Peldszus and Stede~\shortcite{peldszus2015joint} worked on a microtext corpus and aimed to identify the attachment relation between ADUs. Most of the existing research for argumentation mining either focuses on a single task or tackles sub-tasks separately without considering the relation between them. 

%Stab and Gurevych~\cite{stab2014identifying} explored to use supervised approach for both kind of tasks on persuasive essays. Based on the same corpus, Nguyen and Litman~\cite{nguyen2015extracting} employed an external corpus to generate extended topic words ADU type classification in a semi-supervised fashion. Eckle-Kohler et. al.~\cite{TUD-CS-2015180} aimed to use discourse markers to distinguish claims and premises for arguments. 

\begin{figure}
\centering
\includegraphics[width=12cm]{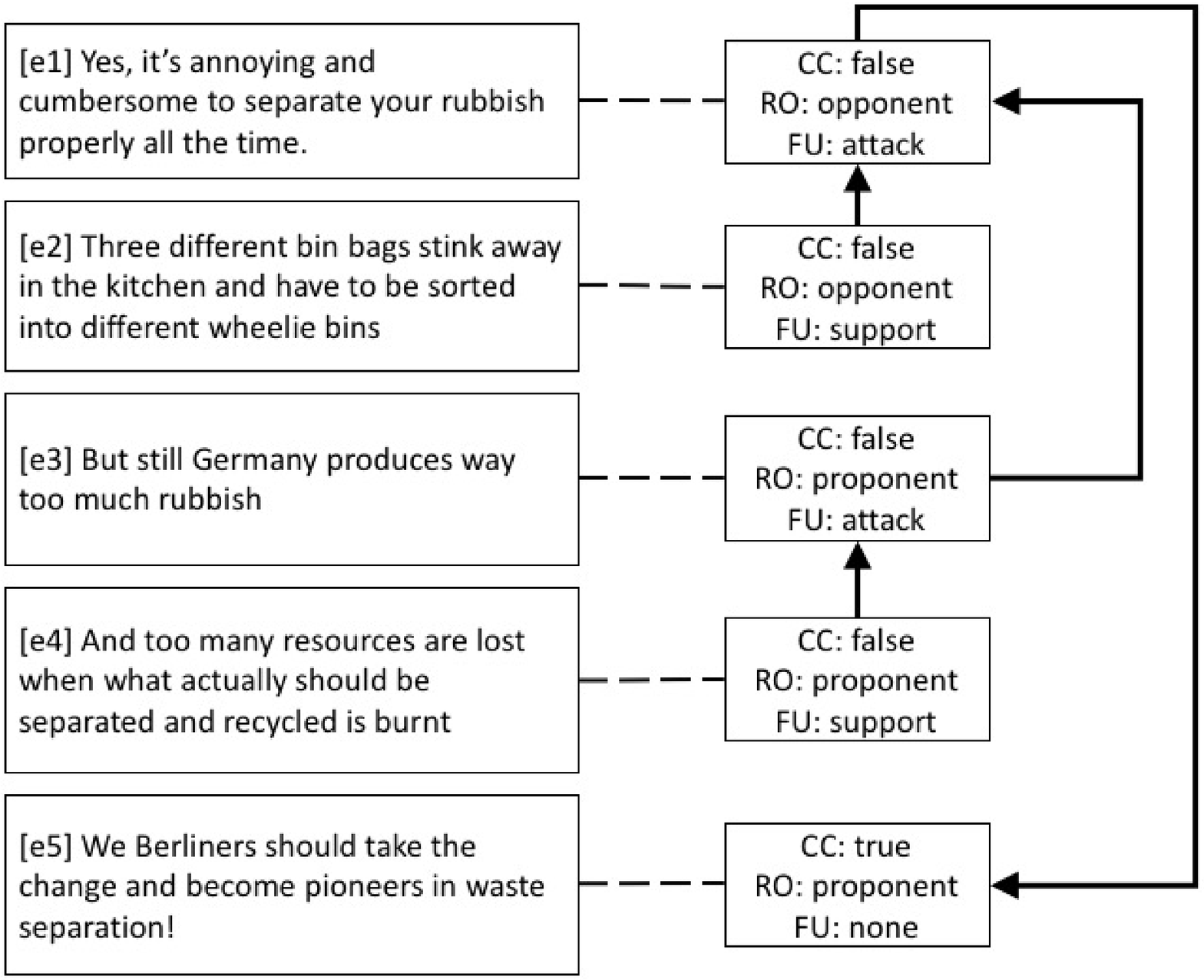}
\caption{An example text and its argumentation structure. Three ADU type labels (CC, RO, FU) and one relation label (arrow): CC stands for central claim; RO stands for role; FU stands for function; arrow stands for attachment relation.}
\label{fig:example}
\end{figure}

Based on the annotation schemes of the argumentative text and characteristics shown in a specific corpus, there can be strong relations between different argumentation sub-tasks. Take the corpus annotated by Peldszus and Stede~\shortcite{peldszus2015annotated} as an example  (Figure~\ref{fig:example} shows an example). There is only one \emph{central claim} in each text unit and there is no attachment relation starting from the \emph{central claim}. It is fair to assume that the result of \emph{central claim} identification is bonded to that of \emph{relation identification}. Stab and Gurevych~\shortcite{stab2014identifying} also showed that if perfect labels from ADU type classification can be added as extra features for relation classification, the performance can be improved significantly, and vice versa. Peldszus and Stede~\shortcite{peldszus2015joint} thus proposed an evidence-graph based approach to jointly solve four argumentation mining sub-tasks. Experiments on a microtext corpus showed its effectiveness. However, their approach requires a tree argumentation structure as input thus is not applicable to many corpora. Especailly, the evaluation corpus is generated artificially, which benefits the joint-model in nature.

In this paper, we also propose to use a joint framework inspired by Roth and Yih~\cite{roth2004conll} to combine the predictions of argumentation mining sub-tasks from separate models for argumentation structure prediction. We treat the problem as an optimization problem. Based on different annotation schemes, we generate corpus-specific constraints and solve the problem using Integer Linear Programming (ILP). With the flexibility of designing corpus-specific objective functions and constraints, our joint model can be applied to all kinds of datasets. We evaluate our model on two public corpora: an artificial generated corpus (microtext) and a real environment corpus (student essays). The experiment results on both corpora show that our joint model improves the performance of separate models significantly. In particular for component-based tasks, our approach can outperform the state-of-the-art joint model based on evidence graph in a large margin.

%The next section describes related work. In Section 3, we briefly introduce our ILP-based joint model. In Section 4 and Section 5, we present details of our approaches and the experiment results on two public corpora. We conclude this paper and point out some future directions in Section 6. 

\section{Related Work}
Previous research on argumentation mining focuses on several sub-tasks, including (1) splitting text into discourse units (DU)~\cite{madnani2012identifying,du2014shell}, (2) identification of ADUs from non-argumentative ones~\cite{moens2007automatic,florou2013argument}, (3) identification of ADU types~\cite{biran2011identifying,nguyen2015extracting,TUD-CS-2015180,TUD-CS-2015178} and (4) identification of relation between ADUs~\cite{lawrence2014mining,stab2014identifying,peldszus2015joint,TUD-CS-2015-0068}. We will concentrate on the latter two sub-tasks in this part and introduce some existing joint models.

\textbf{ADU type classification}: Stab and Gurevych~\shortcite{stab2014identifying} aimed to classify text segments into four classes, namely major claim, claim, premise and non-argumentative for persuasive essays. Based on the same corpus and setting, Nguyen et al.~\shortcite{nguyen2015extracting} explored a semi-supervised method for segment type classification. They proposed to divide words into argument words and topic words; and deploy a semi-supervised method to generate argument words based on 10 seeding words, and then used them as additional features for classification. Habernal and Gurevych~\shortcite{TUD-CS-2015178} also focused on developing semi-supervised features. They exploited clustering of unlabeled data from debate portals based on word embeddings. Some research focused on specific types of ADU identification given its context. Park and Cardie~\shortcite{park2014identifying} proposed to identify the supporting situation for a given claim. Biran et al.~\shortcite{biran2011identifying} aimed to identify justifications for claims. 

%The macro average F-score for all classes is 73\%, the F-score for claim is 54\% and for the major claim is \%63.

\textbf{Relation identification}: There is much less work for relation identification. Palau and Moens~\shortcite{mochales2011argumentation} used a hand-written context-free grammar to predict argumentation trees on legal documents. Kirchner et al.~\shortcite{TUD-CS-2015-0068} presented an annotation study for fine-grained analysis of argumentation structures in scientific publications. For data-driven approaches, Lawrence et al.~\shortcite{lawrence2014mining} constructed tree structures on philosophical texts using unsupervised methods based on topical distance between segments. Stab and Gurevych~\shortcite{stab2014identifying} presented a supervised approach for student essays. Peldszus and Stede~\shortcite{peldszus2015joint} aimed to identify the attachment relation between ADUs on a microtext corpus. %Different from determining the relation for a given segments pair, Wei and Hirst~\shortcite{feng2011classifying} aimed to identify the argument scheme for the given pairs. In particular, they explored to classify argument scheme of the pair of claim and premise into one of the five categories including, from example, from cause to effect, practical reasoning, from consequences and from verbal classification. 

\textbf{Joint model for argumentation mining}: Although there are several approaches for different sub-tasks in argumentation mining, researchers rarely consider to solve sub-tasks in unified way. Stab and Gurevych~\shortcite{stab2014identifying} explored to directly use the prediction results of ADU type classification as features for the task of relation classification, however, without considering logical relation between these two tasks, the effect was marginal. Peldszus and Stede~\shortcite{peldszus2015joint} tackled four argumentation mining tasks including three ADU type classification tasks and the task of relation identification. They proposed to combine the prediction results for these sub-tasks as the edge weights of an evidence graph. They then applied a standard max spanning tree (MST) decoding algorithm and showed its effectiveness on a microtext corpus. 

In our research, we also explore to identify argumentation structure in unified way. We propose to use integer linear programming (ILP) to combine predictions from sub-tasks and generate results jointly with argumentation structure related constraints. Compared to the evidence-graph-based approach that requires the tree-structure of argumentation as input, our model is more flexible and can be easily applied to different corpora with various characteristics.

\section{Framework of ILP-based Joint Model}

\begin{figure}
\centering
\includegraphics[width=12cm]{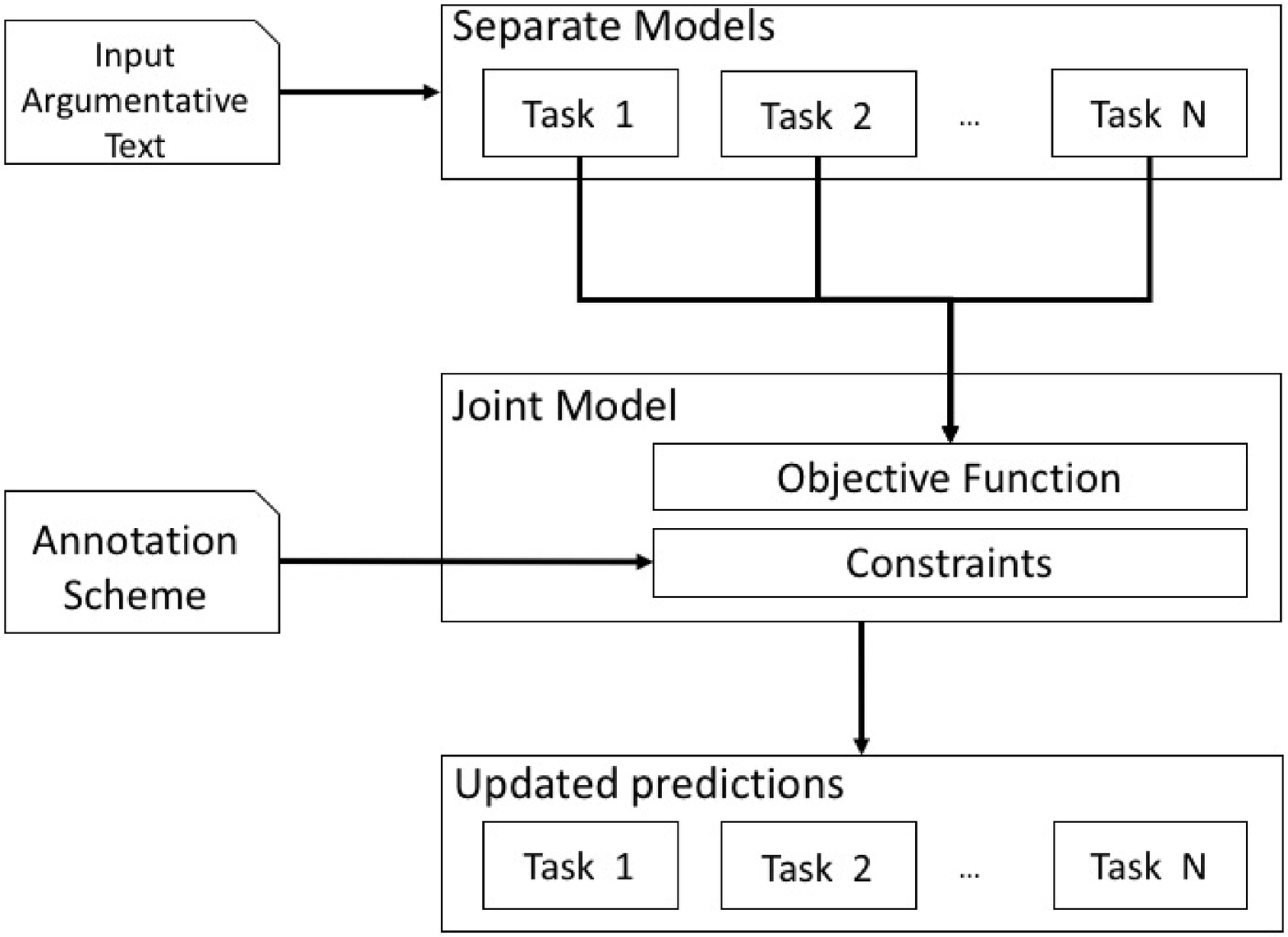}
\caption{ILP-based Joint Model Framework for Argumentation Mining}
\label{fig:framework}
\end{figure} 

The overall framework of our joint model for argumentation mining can be seen in Figure~\ref{fig:framework}. For an input argumentative text unit, we first employ several separate models to predict results for each sub-task. Our joint model then takes the probability scores from each separate model as input to form the objective function. Besides, we define constraints based on the annotation scheme of the target corpus. We solve the objective function by ILP with these constraints. Our joint model finally generates updated predictions for all the sub-tasks simultaneously.  

In order to evaluate our proposed joint model, we use two public corpora, consisting of microtext~\cite{peldszus2015annotated} and persuasive essays~\cite{stab2014annotating} respectively. In the first corpus, all the texts are generated in a controlled environment for research purpose. It ensures the tree structure of argumentations in the input text unit. In the second corpus, all the documents are student essays collected from on online learning platform. It thus contains noise. With different annotation schemes, these two corpora have different sub-task settings. We will design objective functions and construct constraints for each corpus accordingly.

\section{Joint Model for Microtext}

\subsection{Microtext Corpus}

\begin{table}
\begin{center}
\begin{tabular}{c|c|c|c|c|c|c|c|c}
\hline
 \multicolumn{2}{c}{CC}  & \multicolumn{2}{|c}{RO} & \multicolumn{3}{|c}{FU} & \multicolumn{2}{|c}{AT} \\
\hline
true&false&pro&opp&sup&attack&none&at&un-at\\
\hline
112&462&451&125&290&174&112&464&2,000\\
\hline
\end{tabular}
\end{center}
\caption{Statistics of the microtext corpus}
\label{tab:upotsdam-corpus}
\end{table}

The corpus of ``microtexts" is from Peldszus and Stede~\shortcite{peldszus2015annotated}. The dataset contains 112 text units with 576 ADUs. 23 text units are written by the authors, while the rest are generated by instructed users in a controlled environment. The corpus is annotated according to a scheme representing segment-level argumentation structure~\cite{freeman1991dialectics}. The annotations include three ADU type labels, \emph{central claim} (CC, if it is a central claim), \emph{role} (RO, proponent or opponent), \emph{function} (FU, support, attack and none) and one relation label, \emph{attachment} (AT, if there is an attachment relation between a pair of ADUs). An example text unit in this corpus can be seen in Figure~\ref{fig:example}. 

This corpus has the following properties: (1) The length of each text is about 5 ADUs. (2) One segment explicitly states the central claim. (3) Each segment is argumentatively relevant. (4) At least one objection to the central claim is considered. The basic statistics can be seen in Table~\ref{tab:upotsdam-corpus}. There are 2,464 possible pairs of relations between ADUs, in which 464 are annotated as attachment. The corpus contains both German and English version. We only use the English version in our paper. 

%\subsection{Separate Models}

%We implemented the approach of~\cite{peldszus2015joint} for each sub-task\footnote{We obtained similar performance for all the sub-tasks except function classification.}. For ADU type tasks (\emph{cc}, \emph{ro} and \emph{fu}), we extracted language-based features, including binary features for lemma, POS-tags, lemma- and POS-tag-based dependency-parse triples and discourse connectives. Besides, we also extracted some structural features, e.g., segment position, segment length and punctuation count. For relation task, \emph{at}, we combined features extracted for each single segment and added relative distance and their position order as features. We employed MaxEntropy\footnote{\url{http://sourceforge.net/projects/pocket-crf-1/}} for classification because it shows the best result in our implementation. For the sub-task \emph{cc}, \emph{ro} and \emph{at}, we trained binary classifiers, while a three-way classifier was trained for \emph{fu}. 10-fold cross-validation is used to obtain prediction for all the samples.

\subsection{ILP-based Joint Model}

There are four sub-tasks designed in this corpus based on its annotation scheme, including central claim identification (\emph{cc}), role identification (\emph{ro}), function classification (\emph{fu}), and attachment relation classification (\emph{at}). Our joint model takes the probability scores predicted by the individual classifiers for each sub-task as input and generates the final prediction jointly. In order to consider all the sub-tasks simultaneously, we aim to maximize the following objective function:

\vspace{-10pt}

\begin{align}
 w_1 \sum_{i}\{a_{i} CC_{i}\} + w_2 \sum_{i}\{b_{i} RO_{i}\} +  w_3 \sum_{i}\{c_{i} SUP_{i} + e_{i} ATT_{i} + g_{i} NONE_{i}\} + w_4 \sum_{ij}\{d_{ij} AT_{ij}\} \nonumber
\end{align}

\vspace{-5pt}

 %\label{micro_ob}
 
where the four different components correspond to four sub-tasks respectively: $CC_{i}$ stands for the probability of segment \emph{i} being central claim;  $RO_{i}$ stands for the probability of $i$ having proponent role; $SUP_{i}$, $ATT_{i}$, and $NONE_{i}$ denote the probability of the function type of $i$ (\emph{support}, \emph{attack} and \emph{none}); and $AT_{ij}$ is the probability of $i$ attaching to \emph{j}. $a_i$, $b_i$, $c_i$, $e_i$ and $g_i$ are binary variables indicating if segment $i$ is predicted as \emph{true} for different sub-tasks. $d_{ij}$ is also a binary variable representing if segment $i$ attaches $j$. $w_1$, $w_2$, $w_3$ and $w_4$ are introduced to balance the contributions of different sub-tasks.
% $a_i$ and $b_i$ are binary variables indicating that if the node $i$ is a central claim, obtaining proponent role. $c_i$, $e_i$ and $g_i$ are binary variable representing if the node {i}'s attitude to its target node is \emph{support}, \emph{attack} or \emph{none}. 

Based on the task definition and annotation scheme, we have the following constraints.

\begin{enumerate}[a)]
\itemsep-0.2em
	\item There is only one central claim. (Eq.~\ref{upots_st1})
	\item There is at least one opponent segment. (Eq.~\ref{upots_st2})
	\item The function of $i$ is one of \emph{support}, \emph{attack} and \emph{none}. (Eq.~\ref{upots_st7}).
	\item If $i$ is the central claim, then its role is \emph{proponent} (Eq.~\ref{upots_st5}).
	\item If $i$ is the central claim, then its function is \emph{none} (Eq.~\ref{upots_st6}).
	\item If $i$ is the central claim, then it attaches to no other segments, otherwise, it attaches to one and only one other segment. (Eq.~\ref{upots_st8}).
	\item If $i$ is the central claim, it has at least one supporting segment. (Eq.~\ref{upots_st9}).
	\item There can be no more than one relation between two segments. (Eq.~\ref{upots_st10}).
	\item If $i$ attaches $j$ and the function of $i$ is \emph{support}, then the role of $i$ and $j$ are the same. (Eq.~\ref{upots_st3}).
	\item If $i$ attaches $j$ and the function of $i$ is \emph{attack}, then the role of $i$ and $j$ are opposite. (Eq.~\ref{upots_st4}). 

\end{enumerate}

\vspace{-15pt}

\begin{tabular}{rl}
\parbox[t][0.5cm][t]{6cm}{\begin{align}  \sum_{i}a_{i} = 1 \label{upots_st1} \end{align}}& \parbox[t][0.5cm][t]{6cm}{\begin{align}  \sum_{i}\{1 - b_{i}\} >=1 \label{upots_st2}\end{align}  }\\
\parbox[t][0.5cm][t]{6cm}{\begin{align} \forall_{i} c_i + e_i + g_i = 1 \label{upots_st7} \end{align}}& \parbox[t][0.5cm][t]{6cm}{\begin{align} \forall_i a_i <= b_i \label{upots_st5} \end{align}}\\
\parbox[t][0.5cm][t]{6cm}{\begin{align}  \forall_{i} a_i = g_i \label{upots_st6}\end{align}}& \parbox[t][0.5cm][t]{6cm}{\begin{align}  \forall_{i} a_i + \sum_{j} d_{ij} = 1\label{upots_st8} \end{align}}\\
\parbox[t][0.5cm][t]{6cm}{\begin{align} \forall_{j} a_{j} \leq \sum_{i} c_{ij} \label{upots_st9} \end{align}}& \parbox[t][0.5cm][t]{6cm}{\begin{align}  \forall_{ij} d_{ij} + d_{ji} <= 1\label{upots_st10} \end{align}}\\
\parbox[t][1cm][t]{6cm}{\begin{align}  \forall_{ij} (d_{ij} \wedge c_i) \rightarrow (b_i = b_j )\label{upots_st3}\end{align}}& \parbox[t][1cm][t]{6cm}{\begin{align}  \forall_{ij} (d_{ij} \wedge e_i) \rightarrow (b_i + b_j = 1) \label{upots_st4} \end{align}}\\

\end{tabular}

%\begin{align}
%& \sum_{i}a_{i} = 1 \label{upots_st1}\\ 
%& \sum_{i}\{1 - b_{i}\} >=1 \label{upots_st2}\\
%& \forall_{i} c_i + e_i + g_i = 1 \label{upots_st7}\\
%& \forall_i a_i <= b_i \label{upots_st5}\\
%& \forall_{i} a_i = g_i \label{upots_st6}\\ 
%& \forall_{i} a_i + \sum_{j} d_{ij} = 1\label{upots_st8}\\
%&\forall_{j} a_{j} \leq \sum_{i} c_{ij} \label{upots_st9}\\
%& \forall_{ij} d_{ij} + d_{ji} <= 1\label{upots_st10}\\
%& \forall_{ij} (d_{ij} \wedge c_i) \rightarrow (b_i = b_j )\label{upots_st3}\\
%& \forall_{ij} (d_{ij} \wedge e_i) \rightarrow (b_i + b_j = 1) \label{upots_st4}
%\end{align}

\subsection{Results}

We implemented the approach of~\cite{peldszus2015joint} for each sub-task\footnote{We obtained similar performance for all the sub-tasks except function classification.}. For the sub-task \emph{cc}, \emph{ro} and \emph{at}, we trained binary classifiers, while a three-way classifier was trained for \emph{fu}. 10-fold cross-validation is used. We report F-1 score for each sub-task and their macro-F1. We compared three approaches on this corpus. 

%All the probability distributions are normalized before feeding to our joint model. 
\vspace{-5pt}
\begin{enumerate}[-]
\itemsep -0.2em
	\item \textbf{separate}: the baseline approach from~\cite{peldszus2015joint} that we re-implemented.
	\item \textbf{MST}: we implemented the evidence-graph-based approach of~\cite{peldszus2015joint}. In this approach, a fully connected multigraph over all the segments is first generated. Then, the prediction probabilities from different sub-tasks are combined as edge weights. Finally, MST is used to decode the graph and generate a tree structure as argumentation structure. The prediction results for each sub-task are generated from the resulting tree based on some rules. 
	\item  \textbf{ILP}: this is our approach.  
\end{enumerate}

\subsubsection{Overall Experiment Results}
\begin{table}[t]
\begin{center}
\begin{tabular}{l|c|c|c|c|c}
\hline
& cc & ro & fu & at & macro F1\\
\hline
separate & 0.804 & 0.667 & 0.666 & 0.652 & 0.697\\
\hline
MST &\textbf{\underline{0.834}} & 0.686& 0.662&\textbf{\underline{0.696}}&\underline{0.719}\\ 
\hline
ILP &\textbf{\underline{0.834}} & \textbf{\underline{0.695}}& \textbf{\emph{0.681}}&\textbf{\underline{0.696}}&\textbf{\underline{0.727}}\\ 
%\hline
%\hline
%MST$_{tuned}$ & 0.823 & 0.673& 0.666&0.683&0.715\\ 
%\hline
%ILP$_{tuned}$  &0.824 &\textbf{\underline{0.735}}& 0.670& 0.693&\textbf{0.730}\\
\hline
%\hline
%MST-best& \underline{0.845}&\underline{0.702}&0.674&\underline{\textbf{0.704}}&\underline{0.727}\\ 
%ILP-best & \underline{\textbf{0.845}} & \emph{\underline{\textbf{0.737}}} & \emph{\underline{\textbf{0.686}}} & \underline{0.703} & \emph{\underline{\textbf{0.738}}}\\
%\hline
%\hline
%separate(r) & 0.817 & \textbf{0.750} & 0.671 & 0.663&0.725\\
%\hline
%MST(r) & \textbf{0.869} & 0.720 & \textbf{0.710} & 0.693 & \textbf{0.748}\\
%\hline
\end{tabular}
\end{center}
\caption{Performance on microtext corpus (\textbf{Bold}: the best performance in each column; \underline{Underline}: the performance is statistically significantly better than \emph{separate} baseline (p\textless0.05); \emph{Italic}: the performance is significantly better than \emph{MST} (p\textless0.05).)}
\label{tab:upots-overall-performance}
\end{table}

%We also report the best performance for each sub-task on the corpus by varying the combination weight, noted as \emph{MST-best} and \emph{ILP-best}. 

Table~\ref{tab:upots-overall-performance} shows the results on the microtext corpus. \emph{MST} and \emph{ILP} use equal weighting\footnote{We also explored to tune the weighting for each sub-task based on the setting of 10-fold cross validation, however, the performance drops slightly for both joint approaches. The influence of different combination of weighting to the performance of each task will be given in next section.} for sub-task combination.

%We also report the tuned performance for each sub-task on the corpus by varying the combination weight, noted as \emph{MST-tuned} and \emph{ILP-tuned}. The weighting is optimized by 10-fold cross validation based on the performance of each task following \cite{peldszus2015joint}. 

Both joint models (\emph{MST} and \emph{ILP}) can improve performance over \emph{separate} baseline for all the sub-tasks (except \emph{MST} on \emph{fu}) and the improvements are significant\footnote{The significance test is performed using paired two-tailed t-test.} in terms of macro F1. This indicates that joint prediction of sub-tasks can improve the performance of each single task for argumentation mining.
%	\item Compared to \emph{MST}, our approach \emph{ILP} achieves equal or better results for all the sub-tasks. \emph{ILP-best} generates best results for all the sub-tasks (except) and yields significantly better performance than \emph{MST} in terms of macro F1. This shows that our ILP-based approach can better utilize joint information for argumentation mining than the MST-based method. 

Compared to \emph{MST}, our approach \emph{ILP} achieves equal or better results for all the sub-tasks. In particular,  ILP-based approach performs similarly as MST-based model for structure-based tasks (\emph{cc}\footnote{The prediction of major claim can be inferred using relation information.} and \emph{at}), but performs better for component-based tasks (\emph{ro} and \emph{fu}). For \emph{fu}, \emph{ILP} is significantly better than \emph{MST}. For \emph{ro}, \emph{ILP} improve the performance significantly based on \emph{separate} while the improvement obtained by {MST} is not significant. This is partly because we include the probability of function type \emph{none} in the objective function, while the MST-based model could not encode this factor into the graph. Besides, we explicitly state that the number of opponent segments should be at least one, while the MST-based method is unable to set such a constraint. This shows that our ILP-based approach can better utilize joint information for argumentation mining than the MST-based method. 
%\end{enumerate}

\subsubsection{Discussion}
\label{sec:discussion}

\begin{figure*}[t]
\centering
\begin{tabular}{cccc}
\subfloat[cc\label{subfig:cc}]{\includegraphics[width=3.5cm]{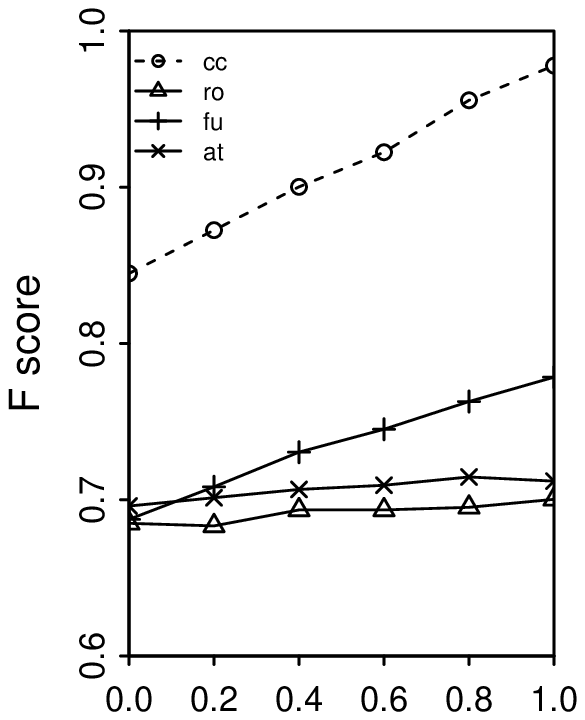}}&
\subfloat[ro\label{subfig:ro}]{\includegraphics[width=3.5cm]{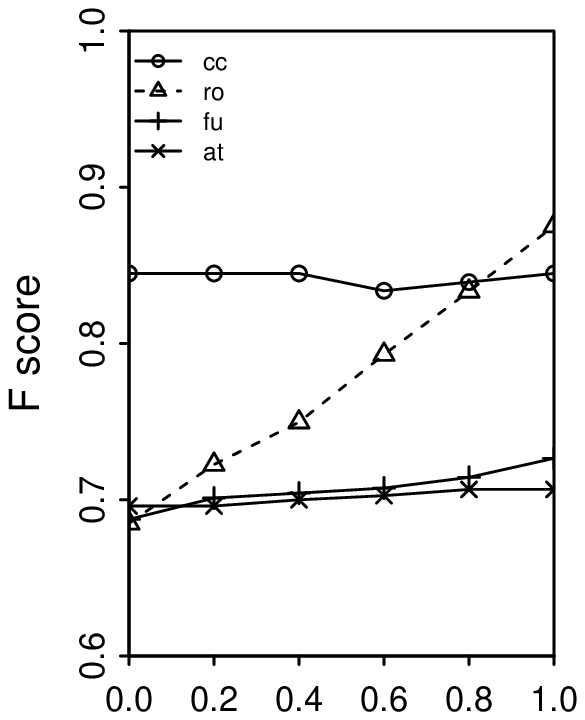}}&
\subfloat[fu\label{subfig:fu}]{\includegraphics[width=3.5cm]{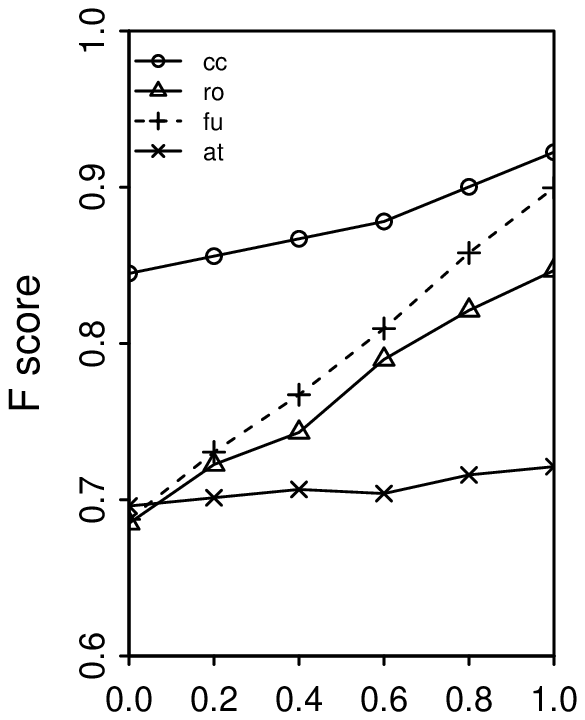}}&
\subfloat[at\label{subfig:at}]{\includegraphics[width=3.5cm]{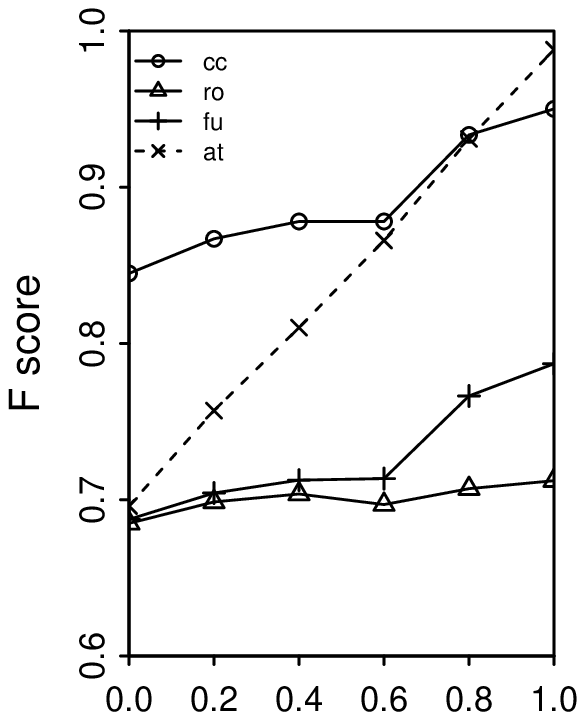}}
\end{tabular}
\caption{Simulations of the effect of better base classifiers in the ILP model using the microtext corpus (x = percentages of predictions overwritten with ground truth; y = F score of each task)}
\label{fig:upotsdam-base-increase}
\end{figure*}

In order to show the contribution of each sub-task to the predictions of the other sub-tasks, we simulate better individual classification results and study their impact. We utilize the strategy proposed in~\cite{peldszus2015joint} for this experiment. We artificially improved the classification result of one sub-task by overwriting a percentage of its predictions with ground truth. The overwritten samples are chosen randomly and the choosing process is on top of the original base classifier, regardless of whether the base classifier already chose the correct label. The simulation result can be seen in Figure~\ref{fig:upotsdam-base-increase}. The figure plots the F score on y-axis for all the sub-tasks when varying the improvement percentage on top of the base classifier for one task (x-axis). Each subfigure shows the simulation results when improving the corresponding task. 

As we can see, function classification is improved when using better role classification due to the logical connection between them, whereas the other sub-tasks are unaffected. Similarly, the performance of role classification is affected by the artificial improvement of function as well. Since the task of central claim and function classification overlaps partially (segment with function \emph{none} is \emph{central claim}), their performance is bonded in this experiment. The performance of attachment classification is difficult to be affected by other sub-tasks, however, improving it results in improvement of other sub-tasks. This is because given a correct tree-structure, the result of other sub-tasks can be inferred to some extent. 

\begin{figure*}
\centering
\begin{tabular}{cccc}
\subfloat[cc\label{subfig:cc}]{\includegraphics[width=3.5cm]{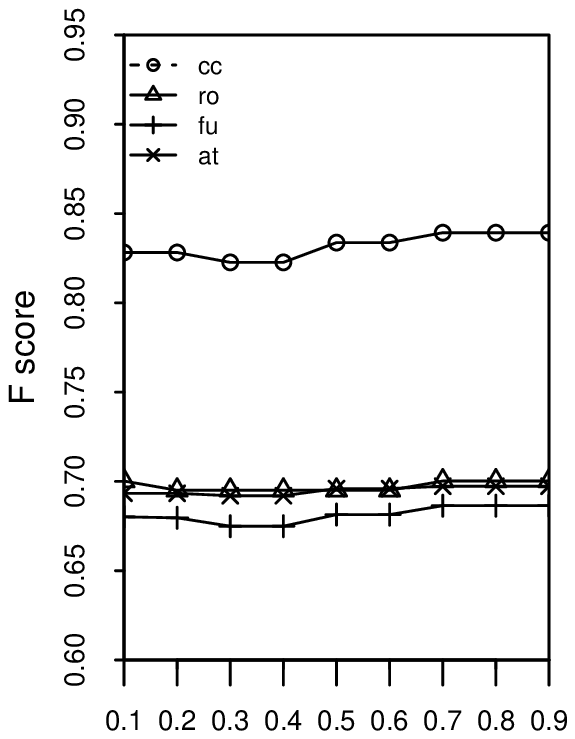}}&
\subfloat[ro\label{subfig:ro}]{\includegraphics[width=3.5cm]{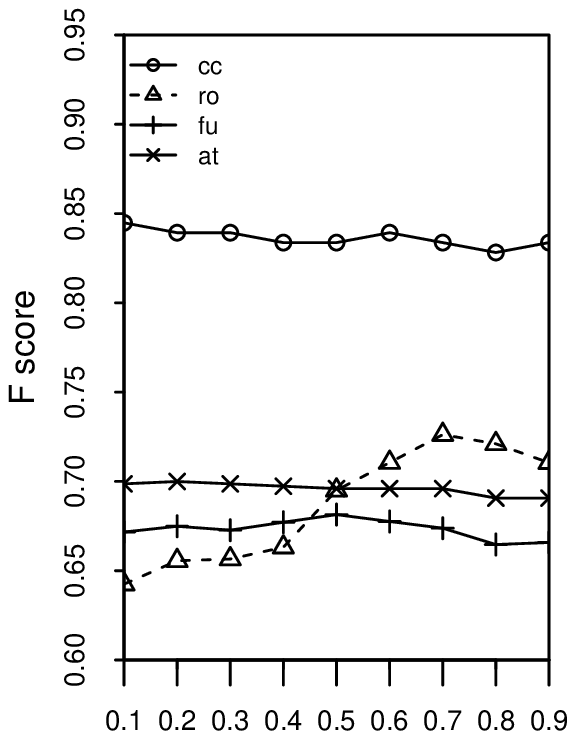}}&
\subfloat[fu\label{subfig:fu}]{\includegraphics[width=3.5cm]{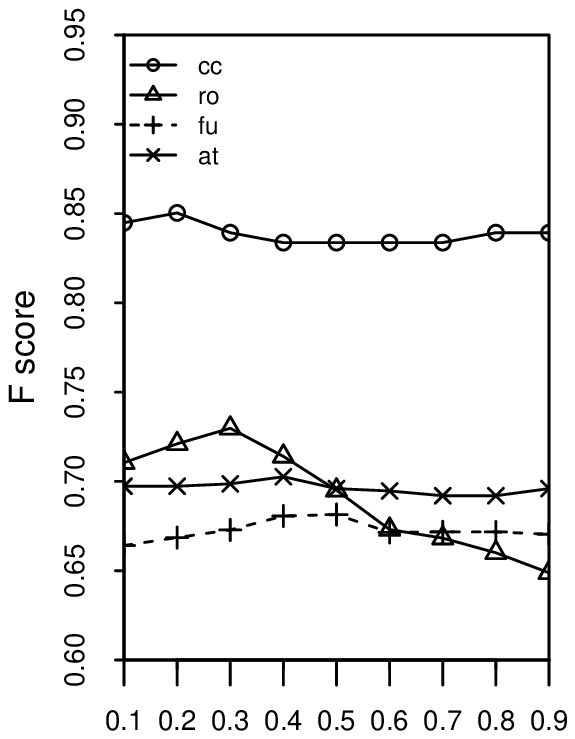}}&
\subfloat[at\label{subfig:at}]{\includegraphics[width=3.5cm]{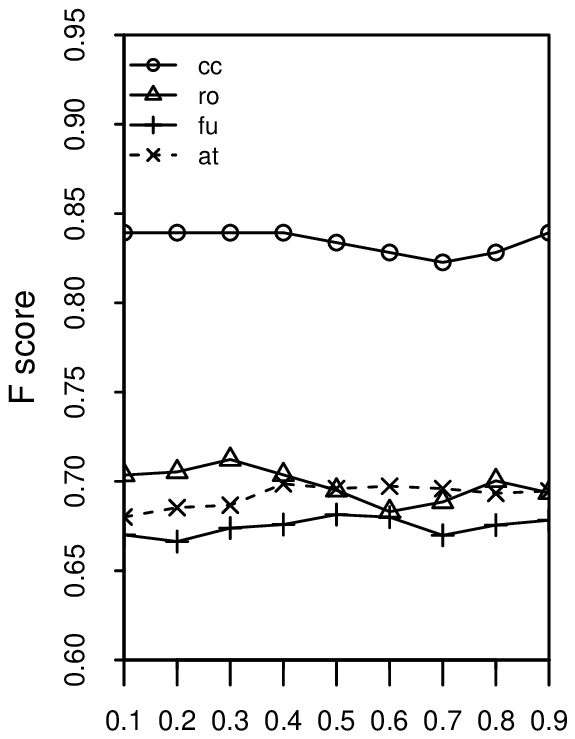}}
\end{tabular}
\caption{The effect of different combination weights using microtext corpus (x = weight for the target task and the weight for each of other 3 tasks is (1 - x) / 3; y = F score of each task)}
\label{fig:upotsdam-weighting}
\end{figure*}

We also evaluate the influence of different combination weights. We set a task as a target at a time, and use $x$ for its weight and other three tasks as ($1 - x$) / 3. The higher $x$ is, the more contribution the target task makes in the joint model. When $x$ equals 0, the label for the target task is assigned randomly, restricted by the constraints only. When $x$ equals 1, the labels for the other three tasks are assigned randomly. We thus sampled $x$ in the range of $[0.1, 0.9]$. The result can be seen in Figure.~\ref{fig:upotsdam-weighting}. In general the performance of the target task increases when its weight increases, although the improvement is not large. \emph{ro} is the only task that is sensitive to different combinations of weight. When the target task is \emph{fu} or \emph{at}, the performance of \emph{ro} drops along with its weight drops.

\section{Joint Model for Student Essays}
The corpus of microtext~\cite{peldszus2015joint} shows the effectiveness of the joint models for argumentation mining, including the evidence-graph-based method as well as our ILP-based approach. However, the corpus is generated for research purpose and follows some restricted rules (e.g., one central claim, one out-arrow for non-central claim, etc.), which might benefit the joint models. In order to show the flexibility of our ILP-based approach for argumentation mining, we applied our ILP-based model on another corpus~\cite{stab2014annotating} consisting of student essays. 

\subsection{Student Essays Corpus}

\begin{table}
\begin{center}
\begin{tabular}{l||c||c|c|c}
\hline
Dataset & Ori & mod-1 & mod-2 & mod-3\\
\hline
para \#&	416 & 351&	288&	233\\
\hline
total comp \#	& 1,552 &1,508&	1,282&	1,077\\
%MajorClaim #	58	48	28
Premise \#&1,033	&1,033&	804&	739\\
%NonArgument #	92	79	64
%Claim #	417	350	246
Claim \#& 519	&475	&398	&274\\
%Claim\# / para &1.25&1.35&1.38&1.18\\
%Premise\# / para &2.48&2.94&2.79&3.17\\
\hline
total rel \#&6,358	&6,358	&4,817	&4,183\\
support rel \#	&989	 &989	&847	&780\\
%Pre $\rightarrow$ Pre \# &	101&	101&	80&	79\\
%Pre $\rightarrow$ Cla \#&	884&	884&	764&	700\\
%Claim\_Claim \# &4 &	4&	3&	1\\
\hline			
\end{tabular}
\caption{Statistics of the student essays corpus}
\label{tab:stab2014corpus}
\end{center}
\end{table}

\vspace{-5pt}

This corpus contains 90 persuasive essays in English. They were selected from the online learning platform essayforum\footnote{\url{http://www.essayforum.com}}. It has segment-level annotations including ADU type annotations of \emph{major claim}, \emph{claim} and \emph{premises}; relations of \emph{support} and \emph{non-support} between argumentative segments. The natural unit of this corpus is an entire student essay, however, there rarely exist cross-paragraph support relations. We thus treat each paragraph as a single unit following~\cite{stab2014identifying}. The original corpus contains 416 paragraphs. 

In the original setting of the corpus, there are two sub-tasks, namely, component classification and relation identification. The former one aims to label a target component as \emph{major claim}, \emph{claim}, \emph{premise} or \emph{non-argumentative}. For a given pair of argumentative components, relation identification aims to classify the relation as \emph{support} or \emph{non-support}. To make the two tasks consistent, we made some modifications on the original corpus. First, we ignored \emph{non-argumentative} components in the component classification task because only relations between argumentative components were considered in relation identification. Second, we combined two categories \emph{major claim} and \emph{claim}, because these two roles are similar when using paragraphs as units, considering there is no claim to claim relation. Third, we filtered those paragraphs with less than one argumentative component because there is no relation identified in such paragraphs. After the pre-selection, we obtained a version of corpus \emph{mod-1}, including 351 paragraphs (versus 416 in the original one). For this modified version of the corpus, the component classification task thus becomes binary classification of \emph{claim} and \emph{premise}.  

In theory, a complete argument should include one claim and its supporting premises. But some essays in the corpus were not written properly. They include isolated claims or premises. To simulate the perfect situation, we further constructed two other versions of the corpus, named \emph{mod-2} and \emph{mod-3}. In \emph{mod-2}, a premise should have at least one out-going arrow to support other components. Furthermore, a claim should have at least one support premise in \emph{mod-3}. The basic statistics of the corpora are shown in Table \ref{tab:stab2014corpus}. Based on the distribution of \emph{claim} and \emph{premise}, this corpus is different from the microtext one in that the number of central claims in each paragraph can be more than one and a premise can support more than one claim. Thus, the tree structure of argumentation is not always assured in this corpus.

\subsection{ILP-based Joint Model}

There are two sub-tasks designed in this corpus based on its annotation scheme, including component classification (\emph{comp}) and relation classification (\emph{rel}). Our ILP model takes the output of the separate models for the two sub-tasks as input and aims to generate results for both tasks in a mutual-reinforcement way. Suppose there are N argumentative components in a target paragraph. For each component $i$, we have a probability distribution for being claim and premise, noted as $P_i$ and $C_i$. For each pair of such components, the probability for component \emph{i} to support \emph{j} is $SUP_{ij}$. We aim to maximize the following objective function: 

\vspace{-10pt}

\begin{align}
v \sum_{i}\{a_{i}C_i + b_{i}P_i\} + (1 - v) \sum_{i,j}\{c_{ij}SUP_{ij}\} \nonumber 
\end{align}

%\label{g_obj}
where $a_i$ and $b_i$ are two binary variables to denote if the target component is claim or premise, $c_{ij}$ is a binary variable to denote if the supporting relation between $i$ and $j$ exists or not. $v$\footnote{Due to the length limit, we will not discuss the influence of $v$ on the student essays corpus.} is introduced to balance the contributions of the two tasks. 

The following constraints are used in this corpus (constraints \emph{f} and \emph{g} are only used in the dataset of \emph{mod-2} or \emph{mod-3}). 

\begin{enumerate}[a)]
\itemsep -0.2em
	\item A component is either claim or premise. (Eq.~\ref{g_st1})
	\item There is at most one relation between two segments. (Eq.~\ref{g_st2})
	\item A relation only starts from a premise. (Eq.~\ref{g_st3}) 
	\item There is at least one claim. (Eq.~\ref{g_st7})
	\item To control the predicted number of support relations, we set the max number of relations in each paragraph as the number of components, $comp\_num$. (Eq.~\ref{g_st8})
	\item A premise should support at least one claim. (Eq.~\ref{g_st5}, for mod-2 and mod-3)
	\item A claim should be supported by at least one premise. (Eq.~\ref{g_st6}, for mod-3)
\end{enumerate}

\vspace{-15pt}

\begin{tabular}{rl}
\parbox[t][0.5cm][t]{6cm}{\begin{align}  \forall_{i}\{a_{i} + b_{i} \} = 1 \label{g_st1} \end{align}}& \parbox[t][0.5cm][t]{6cm}{\begin{align}  \forall_{i,j}\{c_{ij} + c_{ji} \} <= 1 \label{g_st2}\end{align}  }\\
\parbox[t][0.5cm][t]{6cm}{\begin{align} \forall_{j} b_{i} \geq c_{ij} \label{g_st3}\end{align}}& \parbox[t][0.5cm][t]{6cm}{\begin{align}  \sum_{j}\{a_{j}\} >= 1 \label{g_st7} \end{align}}\\
\parbox[t][0.5cm][t]{6cm}{\begin{align}  \sum_{i,j}\{c_{ij}\} <= comp\_num \label{g_st8} \end{align}}& \parbox[t][0.5cm][t]{6cm}{\begin{align}  \forall_{i} b_{i} \leq \sum_{j} c_{ij}\label{g_st5}\end{align}}\\
\parbox[t][1.5cm][t]{6cm}{\begin{align}  \forall_{j} a_{j} \leq \sum_{i} c_{ij} \label{g_st6}\end{align}}&
\end{tabular}

%\begin{align}
%&\forall_{i}\{a_{i} + b_{i} \} = 1  \label{g_st1}\\
%& \forall_{i,j}\{c_{ij} + c_{ji} \} <= 1  \label{g_st2}\\
%&\forall_{j} b_{i} \geq c_{ij} \label{g_st3}\\
%&\sum_{j}\{a_{j}\} >= 1 \label{g_st7}\\
%&\sum_{i,j}\{c_{ij}\} <= comp\_num \label{g_st8}\\
%&\forall_{i} b_{i} \leq \sum_{j} c_{ij} \label{g_st5}\\
%&\forall_{j} a_{j} \leq \sum_{i} c_{ij} \label{g_st6}
%\end{align}

\subsection{Results}

\begin{table*}
\begin{center}
\begin{tabular}{l|ccc|ccc|ccc}
\hline
{\multirow{2}{*}{}} & \multicolumn{3}{c|}{mod-1} & \multicolumn{3}{c|}{mod-2}& \multicolumn{3}{c}{mod-3}\\
\cline{2-10}
 & comp & rel & macro & comp & rel & macro & comp & rel & macro\\
\hline
separate & 0.737 & 0.589 & 0.663 & 0.741& 0.590& 0.665&0.737&0.593&0.665\\
\hline
MST& 0.748& \textbf{\underline{0.666}} & \underline{0.707} & 0.751 & \underline{0.675} & \underline{0.713}&\underline{0.788}& \textbf{\underline{0.689}}&\underline{0.739}\\
\hline
ILP& \textbf{\emph{\underline{0.770}}}& \textbf{\underline{0.666}}& \textbf{\underline{0.718}} & \textbf{\emph{\underline{0.778}}} & \textbf{\underline{0.678}}& \textbf{\underline{0.728}}& \textbf{\underline{0.793}}&\underline{0.688}&\textbf{\underline{0.740}}\\
\hline
%\hline
%MST-best& 0.748& \underline{\textbf{0.668}} & \underline{0.707 }& 0.755& \underline{0.677} & \underline{0.716}& \underline{0.788}&\underline{0.690}&\underline{0.739}\\
%ILP-best& \emph{\textbf{\underline{0.773}}}&\textbf{ \underline{0.668}}& \underline{\textbf{0.720}} & \emph{\underline{\textbf{0.786}}}&  \underline{\textbf{0.681}}&  \underline{\textbf{0.734}}&  \underline{\textbf{0.805}}& \underline{\textbf{0.692}}& \underline{\textbf{0.749}}\\
%\hline
\end{tabular}
\end{center}
\caption{Performance on the student essays corpus (comp: component classification; rel: relation classification; \textbf{Bold}: the best performance in column; \underline{Underline}: the performance is significantly better than separate baseline (p\textless0.05); \emph{Italic}: the performance is significantly better than MST (p\textless0.05).)}
\label{tab:ukp-overall-performance}
\end{table*}

We implemented the approach from~\cite{stab2014identifying} for each sub-task. For component classification, we used five categories of features including structural, lexical, syntactic, indicators and contextual. Instead of SVM, we employed MaxEntropy for classification because it shows better result in our implementation. For relation classification, we followed the original paper and employed MaxEntropy as well. 10-fold cross-validation is used. For both tasks, we trained binary classifiers. We take the probabilities predicted by the two classifiers as input to our ILP model. F-1 score is used as evaluation metric. We compared the following three approaches. 

\vspace{-5pt}

\begin{enumerate}[-]
\itemsep -0.2em
	\item \textbf{separate}: the baseline approach from~\cite{stab2014identifying} that we re-implemented.
	\item \textbf{MST}: we implemented the evidence-graph-based approach for this corpus. Since the relation should always start from a $premise$, we compute the edge score $e_{ij}$ from segment $i$ to $j$ as $\beta P_{i} + (1-\beta) SUP_{ij}$, where $\alpha$ is introduced to balance the contributions of two sub-tasks. 
	\item  \textbf{ILP}: this is our approach.  
\end{enumerate}

\vspace{-5pt}

The performance on the student essays can be seen in Table~\ref{tab:ukp-overall-performance}. In all the three versions of datasets, both joint models can improve the performance significantly compared to the \emph{separate} baseline based on macro F1 score. Our approach \emph{ILP} achieves better macro F1 score than \emph{MST} on all the three datasets and generates significantly better results for component classification in both datasets \emph{mod-1} and \emph{mod-2}. This re-confirms that our ILP-based approach can better utilize joint information for argumentation mining compared to the MST-based methods for component-based task. The performance gain from the joint models over the \emph{separate} baseline approach increases from dataset \emph{mod-1} to \emph{mod-3} when the argumentation structure is more strict. This is because the joint model can perform better when the structure information is more correct. The performance gain of our model over MST-based is larger in dataset \emph{mod-1} and \emph{mod-2} compared to \emph{mod-3}. This shows the MST-based model is more sensitive to the quality of the argumentation structure. Our ILP model is more robust.

\begin{figure}[t]
\centering
\begin{tabular}{cccccc}
\subfloat[mod-1\label{subfig:exp-weight}]{\includegraphics[width=4cm]{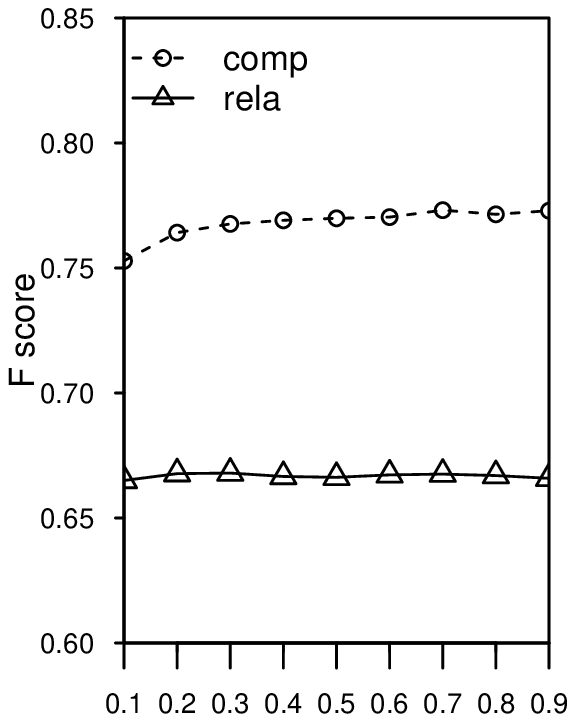}}&
\subfloat[mod-2\label{subfig:res1-weight}]{\includegraphics[width=4cm]{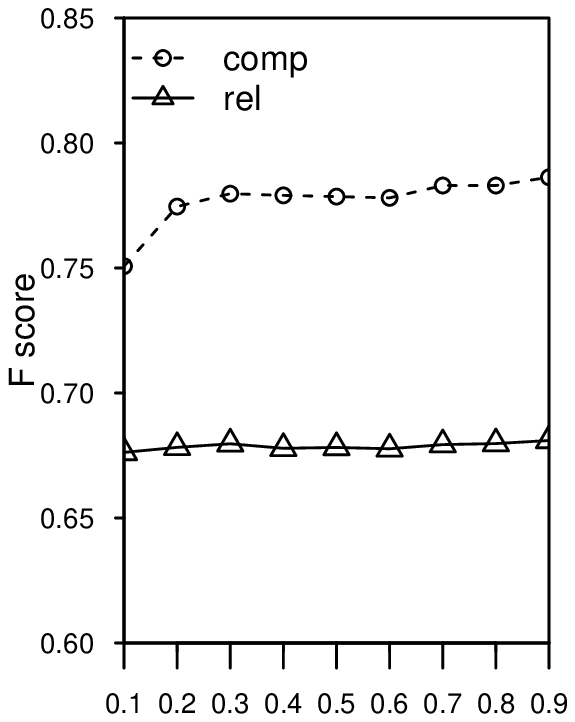}}&
\subfloat[mod-3\label{subfig:res2-weight}]{\includegraphics[width=4cm]{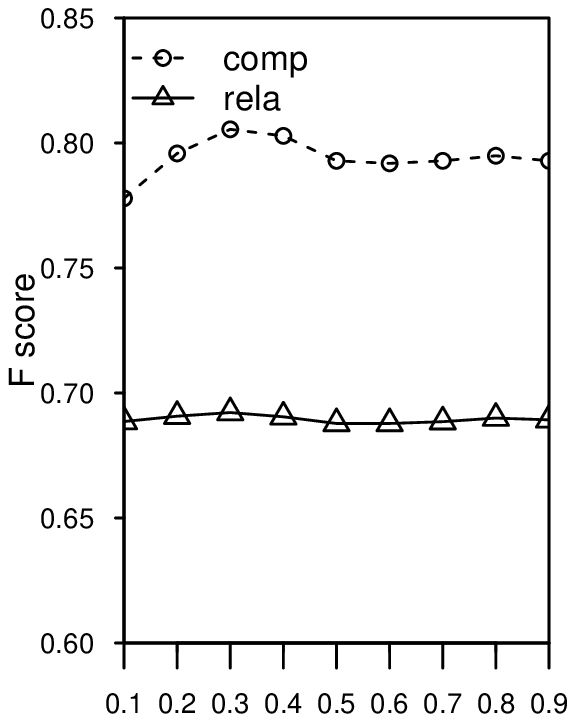}}&
\end{tabular}
\caption{Influence of combination weights using student essays corpus (x-axis = the value of $v$ in Eq.~\ref{g_obj})}
\label{fig:upk-weighting}
\end{figure}

\begin{figure}
\centering
\begin{tabular}{cc}
\subfloat[comp\label{subfig:full-comp}]{\includegraphics[width=6cm]{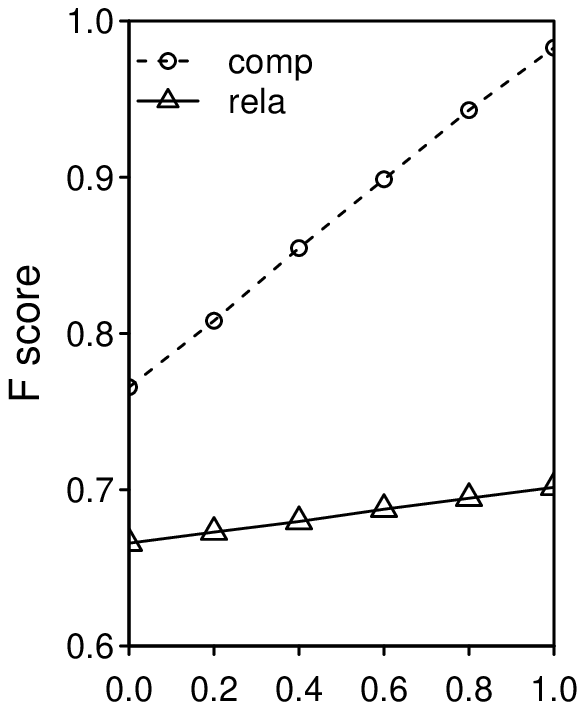}}&
%\subfloat[mod-2 comp\label{subfig:exp-comp}]{\includegraphics[width=2.4cm]{./figures/exp-comp.eps}}&
%\subfloat[mod-3 comp\label{subfig:res-comp}]{\includegraphics[width=2.4cm]{./figures/res-comp.eps}}&
\subfloat[rel\label{subfig:full-rela}]{\includegraphics[width=6cm]{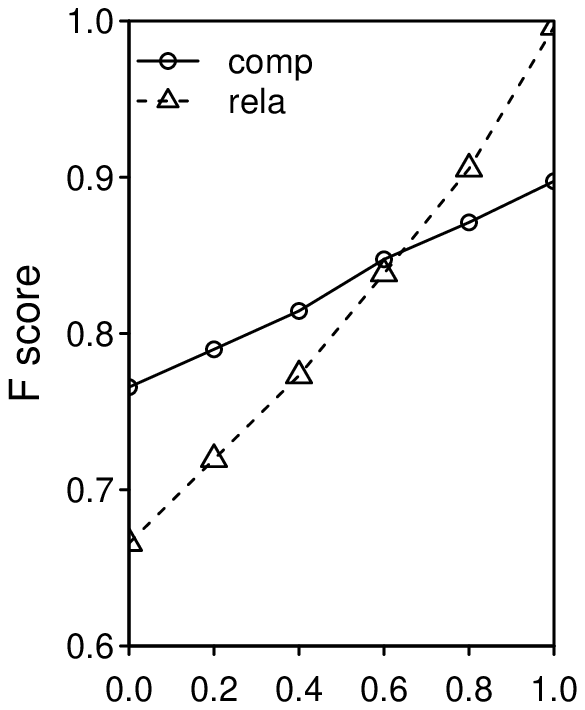}}\\
%\subfloat[mod-2 rel\label{subfig:exp-rela}]{\includegraphics[width=2.4cm]{./figures/exp-rela.eps}}&
%\subfloat[mod-3 rel\label{subfig:res-rela}]{\includegraphics[width=2.4cm]{./figures/res-rela.eps}}
%\subfloat[mod-1\label{subfig:exp-weight}]{\includegraphics[width=2.4cm]{./figures/full-weight.eps}}\\
\end{tabular}
\caption{Simulations of the effect of better base classifiers in the ILP model using \emph{mod-1} (x axis stands for the percentages of predictions overwritten with ground truth; y axis stands for the F score of each task)}
\label{fig:upk-simulation}
\end{figure}

\subsubsection{Discussion}

The result of the simulation experiment on this corpus can be seen in Figure~\ref{fig:upk-simulation}. As we can see, the performance of component classification is greatly improved by a better relation classification. However, the effect of better component classification on relation identification is small. This is because given the correct relation structure of a text, the role of each component can be identified easily. In contrast, even the component classification is perfect, the identification of relation between them is still hard. In addition, the low performance of the base relation identification sub-task also matters.

We also evaluate the effect of different combination of weights in the same way as we did for microtext corpus. The result shows that for larger $\alpha$, the performance for component classification is higher in general; the impact of $\alpha$ on relation classification is smaller. This might be because the base performance for the relation classifier is low. Due to the length limit, we did not show figure here.

The result is shown in Figure~\ref{fig:upk-weighting}. The higher $v$ is, the more contribution component classification makes in the joint model. In general, for larger $v$, the performance for component classification is higher. The impact of $v$ on relation classification is smaller. This might be because the base performance for the relation classifier is low.

\section{Summary and Future Work}

We introduced a joint framework to argumentation mining based on ILP to combine prediction results from individual sub-tasks and encode argumentation structure related characteristics as constraints to generate updated predictions for all sub-tasks. Our ILP-based approach is superior to the existing graph-based model because it can work on all kinds of datasets without requirement for tree structure as input and it can utilize more joint information represented by constraints, especially for component-based tasks. 

There are two interesting directions for future research. First, we will look into other kinds of argumentative texts with more noise and even more complicated argumentation structure, such as posts from online debate forum. Second, we will integrate our argumentation mining framework for automatic student essay scoring.  

%\section*{Acknowledgments}

%Do not number the acknowledgment section. Do not include this section when submitting your paper for review.

\bibliographystyle{acl}
\bibliography{coling2016}

\end{document}